# Calibration of the internal and external parameters of wheeled robot mobile chasses and inertial measurement units based on nonlinear optimization[☆]

Gang Peng[a,b], Zezao Lu[a,b], Zejie Tan[a,b,*], Dingxin He[a,b], Xinde Li[c]

[a] Key Laboratory of Image Processing and Intelligent Control, Ministry of Education;
[b] School of Artificial Intelligence and Automation, Huazhong University of Science and Technology, Wuhan, China;
[c] IEEE senior member, School of Automation, SouthEast University, Nanjing, China

**Abstract**: Mobile robot positioning, mapping, and navigation systems generally employ an inertial measurement unit (IMU) to obtain the acceleration and angular velocity of the robot. However, errors in the internal and external parameters of an IMU arising from defective calibration directly affect the accuracy of robot positioning and pose estimation. While this issue has been addressed by the mature internal reference calibration methods available for IMUs, external reference calibration methods between the IMU and the chassis of a mobile robot are lacking. This study addresses this issue by proposing a novel chassis–IMU internal and external parameter calibration algorithm based on nonlinear optimization, which is designed for robots equipped with cameras, IMUs, and wheel speed odometers, and functions under the premise of accurate calibrations for the internal parameters of the IMU and the internal and external parameters of the camera. All of the internal and external reference calibrations are conducted using the robot's existing equipment without the need for additional calibration aids. The feasibility of the method is verified by its application to a Mecanum wheel omnidirectional mobile platform as an example, as well as suitable for other type chassis of mobile robots. The proposed calibration method is thereby demonstrated to guarantee the accuracy of robot pose estimation.

## 1. Introduction

Mobile robot positioning, mapping, and navigation systems generally adopt a variety of sensors, such as cameras, rangefinder cameras, lidars, odometers, and IMUs, for determining the positions and postures of robots. However, the internal parameters of any sensor cannot be guaranteed to coincide exactly with its design specifications due to manufacturing errors. As such, the internal parameters of sensors must be calibrated prior to use. In addition, the orientation and positioning of each sensor on the robot chassis cannot be guaranteed to be uniform for all of the robots during assembly. Therefore, the external parameters, including the relative position and relative rotation between the individual sensors and between the sensors and robot chassis, must also be calibrated. In addition, the calibration processes employed must be accurate because errors in the internal and external parameters of sensors arising from defective calibration directly affect the accuracy of robot positioning and pose estimation. For example, the external parameters between sensors are used in the equations of robot pose estimation, and the state estimation may fail to converge to an optimal solution due to errors in the external parameters. The calibration of visual odometers represents another example, which requires a detailed

[☆] This work was supported by National Natural Science Foundation of China(No.61672244, No. 91748106), Hubei Province Natural Science Foundation of China(No. 2019CFB526), and Key Technology Project of China Southern Power Grid(GZKJQQ00000164).
  * Corresponding author.
  Gang Peng，PhD, Assoc. Prof, Email: penggang@hust.edu.cn; Zezao Lu(Co-First Author), Master;
  Zejie Tan(Corresponding Author) Master graduate student, Email: tzj@hust.edu.cn; Dingxin He, Master, Prof;
  Xinde Li, PhD, Prof, IEEE senior member.

understanding of the spatial conversion relationships between camera information and wheel speed.

Visual odometers have been extensively applied for mobile robot positioning [1]. This has led to the development of calibration methods focused on the calibration of camera parameters. For example, a specially designed calibration aid has been employed to calibrate the internal and external parameters of cameras [2]. A camera calibration method was proposed to estimate the complete internal and external parameters of an RGB-D camera based on the proscribed motion of a spherical object in front of the camera [3]. In another calibration method, the external parameters between two range cameras can be calibrated by viewing the same plane from different angles. The method relies on the matching of the observation data obtained from the different range cameras [4].

Previous studies have also focused on applying other types of sensors for calibrating the external parameters of cameras for mobile robot positioning, mapping, and navigation. As such, these methods rely on the fusion of data derived from multiple sensors. However, the fusion of data from multiple sensors requires that the external parameters between sensors be known to a high degree of accuracy as well. For example, an automatic calibration method has been proposed, which enables the two-degree-of-freedom calibration of the external parameters of a camera in any state, and this is combined with the information of other external independent motion sources, such as a wheel speed odometer, to complete an accurate six-degree-of-freedom calibration [5]. A calibration method was also proposed for the simultaneous calibration of a camera and wheel speed odometer installed on a differential-driven robot [6]. Here, the complete internal and external parameters of the camera, as well as those of the odometer, can be calibrated based only on the wheel speed and a collection of proscribed camera images captured for a set of known landmarks. Here describes a novel and a low-cost calibration approach to estimate the relative transformation between an IMU and a camera, which are rigidly mounted together [7]. A method for calibrating the external parameters of two-dimensional (2D) lasers and cameras has also been proposed, which relies on the observations of orthogonal trihedrons [8]. However, the above-discussed calibration methods all require the use of prior knowledge regarding the environment or use artificially designed calibration aids. This has been addressed by the development of a two-step analysis approach to complete the external parameter calibration of cameras and odometers based on the least square method [9]. The method requires no initial values or special hardware scenarios. The first step estimates a subset of the external parameters by analyzing the defined least squares problem, and then determines the remaining external parameters by passing these initial parameters as measurement constraints in the defined least squares problem.

As discussed, the above methods were developed for calibrating the external parameters of the cameras employed in mobile robot positioning, mapping, and navigation systems. However, the widespread application of low-cost sensors based on microelectromechanical systems (MEMS) has facilitated the application of inertial measurement units (IMUs) to obtain acceleration and angular velocity measurements for mobile robots. We note that the information derived from IMUs is entirely complementary to that provided by wheel speed odometers and cameras. Accordingly, mobile robot positioning, mapping, and navigation systems can be designed with greatly enhanced robustness and accuracy by combining visual and wheel speed measurements with IMU measurements. However, errors in the internal and external parameters of an IMU arising from defective calibration also affect the accuracy of robot positioning and pose estimation directly. While this issue has been addressed by the mature IMU internal reference calibration methods presently available for IMUs, accurate external parameter calibration methods between the IMU and the chassis of a mobile robot remain poorly developed.

This study addresses this issue by proposing a novel chassis–IMU internal and external parameter calibration algorithm based on nonlinear optimization, which is designed for robots equipped with cameras, IMUs, and wheel speed odometers. All of the internal and external reference calibrations are conducted by the IMU without the need for additional calibration aids. The process of the proposed calibration method is illustrated by the flow chart given in Fig. 1. Here, the process is simplified by decoupling the parameter calibration into two main steps. The first step is employed when robot motion is restricted to the horizontal plane, which limits rotations to occur only about the

Z axis of the chassis. Then, the pitch and roll angles of the IMU in the chassis coordinate system are obtained by calculating the direction of the rotation axis in the IMU coordinate system by principal component analysis (PCA). In the second step, visual inertial odometry (VIO), which is conducted using the VINS-Mono algorithm, and wheel speed inertial odometry are employed to record the pose data of the robot over a specific period [10]. The relationship between the relative poses between successive time frames is employed as pose data with which a nonlinear least squares problem is defined for optimizing both the interior parameters of the chassis and chassis–IMU external parameters. The feasibility of the method is verified by its application to a Mecanum wheel omnidirectional mobile platform equipped with an Intel RealSense ZR300 depth camera, wheel speed odometer, and an IMU as an example. The sensors on the experimental mobile platform are connected by rigid bodies, and the spatial distance between the camera and the IMU is about 5 cm, while the IMU is positioned about 60 cm from the center of the chassis. As such, the spatial distance between the sensors is relatively large, and the same origin cannot be applied to each sensor. The example demonstrates the ease with which the proposed calibration algorithm is implemented, and the process is shown to guarantee the accuracy of calibration.

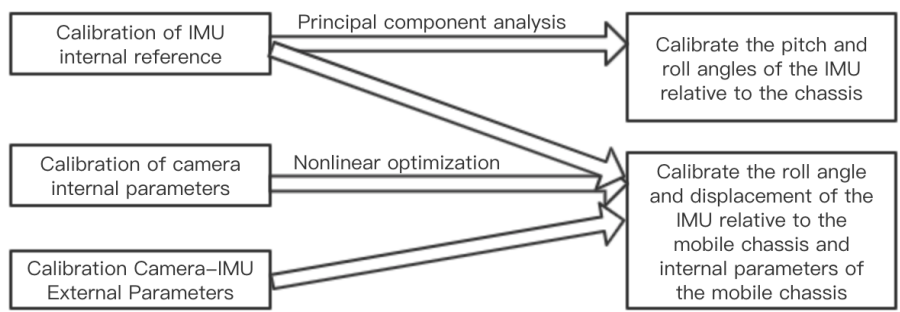

Figure 1. Calibration process

## 2. Coordinate systems and variable definitions

### 2.1 Coordinate systems

As illustrated in Fig. 2, the four coordinate systems of interest in the present work are the IMU coordinate system, which is denoted as the body coordinate system (**B**), the camera coordinate system (**C**), the chassis coordinate system, which is denoted as the wheel odometer coordinate system (**O**), and the intermediate coordinate system, denoted as the fake-body system (**F**), which is employed to simplify calculations. The origins of the **B** and **C** coordinate systems are located at the origins of the respective sensors, while the origin of **O** is the center point between the wheels of the robot chassis and the **F** coordinate system is located on the XY plane of **O** with its origin given as the projection of the origin of **B** on the XY plane of the robot chassis.

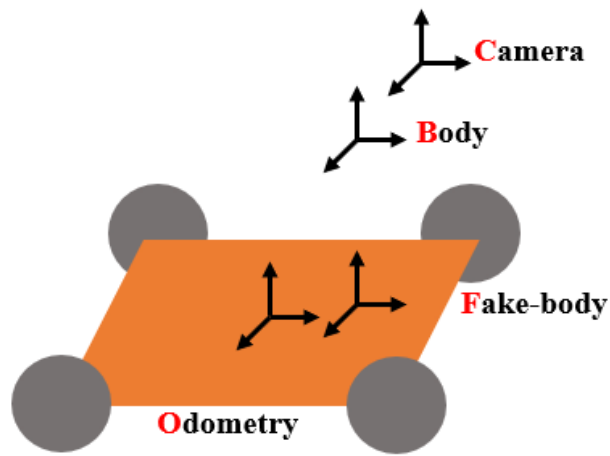

Figure 2. Definitions of coordinate systems for the inertial measurement unit (**B**), camera (**C**), wheel odometer (**O**),

and intermediate coordinate system (**F**).

## 2.2 Variable definitions

Variables within the reference coordinate system of **O** in the *k*-th frame are denoted as $(\cdot)^{O_k}$, while variables within the reference coordinate system of **B** in the *k*-th frame are denoted as $(\cdot)^{B_k}$. The rotation matrix from a generalized **S₂** coordinate system to a generalized **S₁** coordinate system is denoted as $\mathbf{R}_{S_2}^{S_1} \in SO(3)$, where SO(3) is a rotation group belonging to a special orthogonal group, while $\mathbf{p}_{S_2}^{S_1} \in \mathbb{R}^3$ represents the spatial position of the origin of the **S₂** coordinate system in the **S₁** coordinate system and $\mathbf{T}_{S_2}^{S_1} = \begin{bmatrix} \mathbf{R}_{S_2}^{S_1} & \mathbf{p}_{S_2}^{S_1} \\ 0 & 1 \end{bmatrix} \in SE(3)$ represents the transformation matrix from the **S₂** coordinate system to the **S₁** coordinate system, where SE(3) is a special Euclidean group whose elements are denoted as rigid motions or Euclidean motions, and comprise arbitrary combinations of translations and rotations, but not reflections. Finally, terms of the form $\widehat{(\cdot)}$ represent measurement data with noise or estimates for a variable.

Rotations are represented in this study by three possible notations: the rotation matrix $\mathbf{R} \in SO(3)$, rotation at an angle $\boldsymbol{\theta} \in \mathbb{R}^3$ about an axis passing through the origin of the respective coordinate system, and Euler angles, which use three successive rotations about the coordinate axes to represent a rotation in three dimensions. These notations are defined as follows:

(1) Rotation matrix

The rotation matrix has three degrees of freedom, and is defined as $\mathbf{R} \in \mathbb{R}^3$, where the constraints $\mathbf{R}\mathbf{R}^T = \mathbf{I}$, with **I** representing the identity matrix, and the determinant of **R** (i.e., det(**R**)) = 1 must be satisfied.

(2) Shaft angle

The shaft angle of rotation is defined as follows:

$$\boldsymbol{\theta} = \mathbf{u}\phi = \begin{bmatrix} u_x\phi \\ u_y\phi \\ u_z\phi \end{bmatrix} = \begin{bmatrix} \theta_x \\ \theta_y \\ \theta_z \end{bmatrix}, \tag{1}$$

where $\mathbf{u} = [u_x \quad u_y \quad u_z]^T$ represents the rotation axis, which satisfies the constraint det(**u**) = 1, and $\phi$ represents the rotation angle.

(3) Euler angle

Euler angles have various definitions depending on the order of the rotations about the coordinate axes. This article employs the order designated as $Z - Y' - X''$, where $Z$, $Y'$, and $X''$ represent the yaw (or heading), pitch, and roll (*YPR*) angles, respectively, to decompose a three-dimensional (3D) rotation into three rotations about the individual axes. Rotation is first considered about the Z axis at an angle *Y*. This is followed by rotation about the Y axis at an angle *P*, and finally by rotation about the X axis at an angle *R*. The conversion relationship between the YPR Euler angles and **R** is given as follows:

$$\mathbf{R}\left\{\begin{bmatrix} yaw \\ pitch \\ roll \end{bmatrix}\right\}_{YPR} = \mathbf{R}\left\{\begin{bmatrix} 0 \\ 0 \\ yaw \end{bmatrix}\right\} \times \mathbf{R}\left\{\begin{bmatrix} 0 \\ pitch \\ 0 \end{bmatrix}\right\} \times \mathbf{R}\left\{\begin{bmatrix} roll \\ 0 \\ 0 \end{bmatrix}\right\} = \begin{bmatrix} c_1c_2 & c_1s_2s_3 - c_3s_1 & s_1c_3 + c_1c_3s_2 \\ c_2s_1 & c_1c_3 + s_1s_2s_3 & c_3s_1s_2 - c_1s_3 \\ -s_2 & c_2s_3 & c_2c_3 \end{bmatrix}. \tag{2}$$

Here, the following definitions are applied in (2):

$$\begin{cases} c_1 = \cos(yaw), s_1 = \sin(yaw) \\ c_2 = \cos(pitch), s_2 = \sin(pitch). \\ c_3 = \cos(roll), s_3 = \sin(roll) \end{cases}$$

According to (2) and (3) above, the conversion relationship between **R** and the YPR Euler angles can be obtained as follows:

$$\begin{cases} yaw = \operatorname{atan}(R_{12}, R_{11}) \\ pitch = \operatorname{atan}(-R_{13}, R_{11}\cos(yaw) + R_{12}\sin(yaw)) \\ roll = \operatorname{atan}(R_{13}\sin(yaw) - R_{23}\cos(yaw) - R_{12}\sin(yaw) + R_{22}\cos(yaw)) \end{cases}. \tag{3}$$

Here, terms $R_{ij}$ represent the element in the *i*-th row and the *j*-th column of **R**.

## 3. Model analysis

### 3.1 IMU internal parameters

#### 3.1.1 Systematic errors in IMU measurements

For an ideal six-axis IMU, the three axes of the accelerometer and the three axes of the gyroscope, respectively, correspond to the X, Y, and Z axes of a 3D Cartesian coordinate system. Here, an accelerometer measures acceleration along the X, Y, and Z directions, while a gyroscope measures the angular velocity about the X, Y, and Z axes. However, limitations in manufacturing accuracy and other reasons induce deviations from ideal axial alignments in actual IMUs, where the three axes of an accelerometer cannot be perfectly orthogonal, and the three axes of a gyroscope cannot be perfectly aligned with the three axes of the accelerometer. This condition is denoted as axis deviation. The measured acceleration or angular velocity also deviates from the actual acceleration or angular velocity, and the ratio of the measured value to the actual value is denoted as the scale factor. In addition, measurements of the accelerometer and gyroscope under actual conditions of zero acceleration and rotational velocity are typically not zero, and this condition is denoted as zero offset.

The example implementation employed in the present work adopts a BMI055 six-axis IMU (Bosch Sensortec). According to the data sheet description, the typical cross-axis sensitivity of the accelerometer and gyroscope is ±1%, which means that the measurement value for each axis will be affected by the measurement values of the other axes orthogonal to it by ±1%. This error can be compensated by calibrating the IMU according to the axis deviation. In addition, the typical sensitivity tolerance of the gyroscope is ±1%, which means that the ratio between the measured angular velocity and the actual value has an error of about ±1%. This error can be compensated by calibrating the IMU according to the scale factor. Finally, the typical zero-g offset of the accelerometer in the *x*, *y*, and *z* directions is ±70 mg, the typical zero-g offset temperature drift is ±1 mg/K, and the typical zero-g offset supply voltage drift is 0.5 mg/V, which means that the offset error is generally the result of the comprehensive influence of the sensor's internal structure, temperature, and other changes. This error can be compensated by calibrating the IMU according to the zero offset.

An IMU internal calibration method has been developed that requires no external reference device, and can provide calibration results for axis deviation, scale factor, and zero offset [11]. The accelerometer measurement model employed by this method is defined as follows:

$$\mathbf{a}^o = \mathbf{T}_a \mathbf{K}_a (\mathbf{a}^S + \mathbf{b}_a + \mathbf{\eta}_a), \tag{4}$$

where $\mathbf{a}^o$ is the measured value of the accelerometer in the orthogonal coordinate system, $\mathbf{T}_a = \begin{bmatrix} 1 & -\alpha_{yz} & \alpha_{zy} \\ 0 & 1 & -\alpha_{zx} \\ 0 & 0 & 1 \end{bmatrix}$ is the accelerometer axis deviation, $\mathbf{K}_a = \begin{bmatrix} s_x^a & 0 & 0 \\ 0 & s_y^a & 0 \\ 0 & 0 & s_z^a \end{bmatrix}$ is the scale factor, $\mathbf{a}^S$ is the original measured value in the actual coordinate system of the accelerometer, $\mathbf{b}_a = [b_{ax} \quad b_{ay} \quad b_{az}]^T$ is the zero offset, and $\mathbf{\eta}_a$ is additive white noise. The gyroscope measurement model is defined as follows:

$$\mathbf{w}^o = \mathbf{T}_g \mathbf{K}_g (\mathbf{w}^S + \mathbf{b}_g + \mathbf{\eta}_g), \tag{5}$$

where $\mathbf{w}^o$ is the measured value of the gyroscope in the orthogonal coordinate system, $\mathbf{T}_g = \begin{bmatrix} 1 & -\gamma_{yz} & \gamma_{zy} \\ \gamma_{xz} & 1 & -\gamma_{zx} \\ -\gamma_{xy} & \gamma_{yx} & 1 \end{bmatrix}$ is the gyroscope axis deviation, $\mathbf{K}_g = \begin{bmatrix} s_x^g & 0 & 0 \\ 0 & s_y^g & 0 \\ 0 & 0 & s_z^g \end{bmatrix}$ is the scale factor, $\mathbf{w}^S$ is the original measured value in the actual coordinate system of the gyroscope, $\mathbf{b}_a = [b_{gx} \quad b_{gy} \quad b_{gz}]^T$ is the zero offset, and $\mathbf{\eta}_g$ is additive white noise.

### 3.1.2 Random errors in IMU measurements

This article applies the Allan variance method to estimate the random error of the accelerometer and gyroscope. Here, the Allan variance method is a time-domain analysis technique that can determine the noise process of any signal [12]. First, the variance of the original measurement data is calculated under different bandwidth mean filters, and the white noise and bias instability parameters of the signal are obtained by formula fitting to the variation trends of the variance with respect to the filter bandwidth. In the present work, the white noise corresponds to $\mathbf{\eta}_a$ and $\mathbf{\eta}_g$, while the bias instability parameters correspond to $\dot{\mathbf{b}}_a$ and $\dot{\mathbf{b}}_g$.

## 3.2 Projection model of the camera

Conventional lens cameras typically employ the following normalized pinhole camera model:
$$z\mathbf{P}_c \triangleq \mathbf{KP}, \tag{6}$$
where $z$ is the depth of the target, $\mathbf{P}_c = [u \quad v \quad 1]^T$ is the homogeneous normalized pixel coordinates of the target, $\mathbf{K}_c = \begin{bmatrix} f_x & f_x\alpha & c_x \\ 0 & f_y & c_y \\ 0 & 0 & 1 \end{bmatrix}$ is the camera projection matrix, and $\mathbf{P} = \begin{bmatrix} x \\ y \\ z \end{bmatrix}$ is the position of the target in the camera coordinate system. Here, $f_x$ and $f_y$ are the focal lengths along the X and Y axes, respectively, with units of pixels/m, $c_x$ and $c_y$ are the principal points where the X and Y axes intersect the image plane, respectively, which are given in units of pixels, and the parameter $\alpha$ is used to correct non-orthogonality between the X and Y axes. However, the projection model of a camera cannot be fully equivalent to the pinhole camera model given in (6) due to errors during lens fabrication and mounting, as well as design reasons. Therefore, the model must be calibrated to compensate for the radial distortion caused by deviations in the lens and the tangential distortion caused by deviations in the lens mounting.

Radial distortion, which is also denoted as barrel distortion or pincushion distortion, is addressed as follows:
$$\begin{aligned} x_{distorted} &= x(1 + k_1 r^2 + k_2 r^4 + k_3 r^6) \\ y_{distorted} &= y(1 + k_1 r^2 + k_2 r^4 + k_3 r^6). \end{aligned} \tag{7}$$
Here, $(x_{distorted}, y_{distorted})$ are the normalized pixel coordinates after removing distortion, $(x, y)$ are the original normalized pixel coordinates, $r$ is the distance (radius) from a pixel to a main point, and $k_1$, $k_2$, and $k_3$ are radial distortion parameters, where $k_3$ is optional, and is set to 0 when disregarded, as it is in this study. Tangential distortion, which is also denoted as keystone distortion, is addressed as follows:
$$\begin{aligned} x_{distorted} &= x + [2p_1 xy + p_2(r^2 + 2x^2)] \\ y_{distorted} &= y + [p_1(r^2 + 2y^2) + 2p_2 xy]. \end{aligned} \tag{8}$$
Here, $p_1$ and $p_2$ are tangential distortion parameters.

The Intel RealSense ZR300 depth camera includes a standard lens RGB camera with a rolling shutter, and a grayscale fisheye camera with a global shutter. The pinhole camera model presented in (6) is suitable for the RGB camera. However, the fisheye camera involves a degree of distortion that is too severe for applying the pinhole camera model directly. Therefore, we employ the universal camera model [13] to define the camera projection. This model applies a convex mirror in front of a pinhole camera to model a fisheye camera. This adds a parameter $\xi$ to the pinhole camera model in (6) to model the impact of the convex mirror on the camera projection. Here, the camera model approaches the pinhole camera model as the value of $\xi$ approaches 1. The complete camera model parameters used in this paper are therefore $(f_x, f_y, c_x, c_y, k_1, k_2, p_1, p_2, \xi)$.

## 3.3 Mecanum wheel movement model

The Mecanum wheel chassis has three degrees of freedom, and can be rotated and translated with two degrees of freedom simultaneously. Therefore, this wheel is suitable for narrow and complex environments, but it has high requirements for ground quality because Mecanum wheels can lose traction on ground that is soft or uneven, which seriously detracts from the dead reckoning effect of the wheel odometer.

The configuration of the Mecanum wheel chassis used in the present work is illustrated in Fig. 3. The forward direction of the vehicle body is defined as the positive X direction of the chassis, the left side of the vehicle body is defined as the positive Y direction, and the center point between the four wheels is defined as the center of the chassis, which, as discussed, is the origin of the **O** coordinate system.

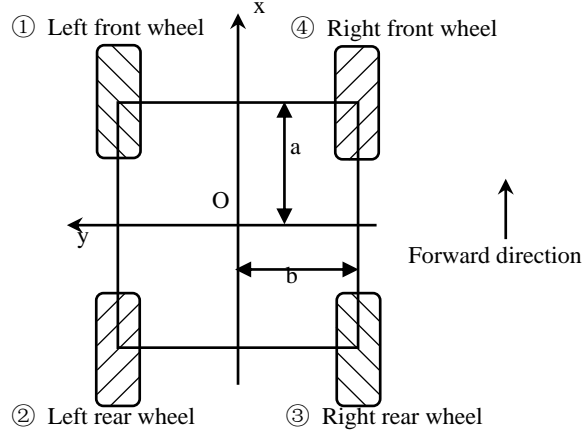

Figure 3. Top view of the Mecanum wheel chassis

Due to installation errors, manufacturing tolerances, and variations in the rolling friction of rubber tires, the actual parameters in the motion model of the Mecanum wheel chassis, such as the wheel radius and wheelbase, deviate from the design parameters, which will affect the dead reckoning accuracy of the wheel odometer, and these parameters must therefore be calibrated. Accordingly, the measured real-time velocity of the chassis $\hat{\mathbf{v}}_{\text{base}}$ is defined as follows:

$$\hat{\mathbf{v}}_{\text{base}} = \mathbf{K}\mathbf{v}_{\text{base}} + \mathbf{\eta}_o, \tag{9}$$

where $\hat{\mathbf{v}}_{\text{base}}$ is the measured real-time velocity of the chassis, $\mathbf{K} = \begin{bmatrix} s_x & 0 & 0 \\ 0 & s_y & 0 \\ 0 & 0 & s_z \end{bmatrix}$ is a diagonal matrix containing the errors $s_x$, $s_y$, and $s_z$ associated with the X axis, Y axis, and the scale factor of the rotation speed, respectively, $\mathbf{v}_{\text{base}}$ is the actual velocity of the chassis, and $\mathbf{\eta}_o$ is the added white noise of the velocity measurements. Here, $\mathbf{K}$ is diagonal because it is assumed that the X and Y axes of the chassis are orthogonal, and errors in the rotational angular velocity and the translational linear velocity are independent of each other.

## 4. Parameter analysis and calibration example

### 4.1 Chassis internal parameters and chassis−IMU external parameters

The motion model of the chassis limits robot motion to a 2D surface. Accordingly, the robot can only generate acceleration excitations in the X and Y directions, and angular velocity excitations in the Z direction. However, robot motion along a curved surface with large angle fluctuations can generate acceleration and angular velocity excitations in three directions. Moreover, angular velocity excitation in at least two directions is required to determine the relative rotation between two coordinate systems on the same rigid body. It is therefore not possible to directly determine the relative rotation between the chassis and the IMU using only angular velocity measurements. This is addressed in the present work by dividing the calibration of external parameters between the chassis and the IMU into two steps involving the calibration of the pitch and roll angles, followed by the calibration of the heading angle and displacement.

#### 4.1.1 Pitch and roll angle calibration based on principal component analysis

When the robot is moving in the horizontal plane, the pitch and roll angles of the IMU in the gravity coordinate system can be determined by the direction of the acceleration due to gravity in the IMU coordinate system. However, the zero offset of the accelerometer induces error in determining the direction of this acceleration. Even after calibration, changes in temperature, deformation, and other reasons will result in a large zero offset uncertainty. The zero offset range of the BMI055 IMU is approximately ±0.686 m/s$^2$, and using this to calibrate the direction of gravity will result in a maximum angular error of $\text{atan}(\frac{0.686}{9.8}) \approx 4°$.

This can be addressed by noting that rotation will occur only about the Z axis of the chassis when the robot is moving in a horizontal plane. Using this prior knowledge, the gyroscope can be used to calibrate the pitch and roll angles of the IMU in the **O** coordinate system if the direction of the rotation axis is calculated in the **B** coordinate system. While the zero offset of the gyroscope will also induce calibration error, this can be quickly determined by holding the robot motionless for a short time, which would provide a gyroscope zero offset of less than 0.1° s$^{-1}$. For example, the zero offset of the gyroscope would produce an angular error of only about 0.2° when employing the gyroscope to determine the direction of rotation with the robot rotating at an angular velocity of 30° s$^{-1}$.

Accordingly, the present work employs the gyroscope to calibrate the pitch and roll angles of the IMU in the **O** coordinate system. To this end, the average axis of rotation for all angular velocity measurements are calculated during robot motion. Here, the rotation axis will be distributed on both sides of the chassis during clockwise and anticlockwise rotations with the center of the chassis as the origin. As such, the average rotation axis cannot be obtained directly by averaging each rotation axis. This is addressed by applying PCA to extract the direction to which the variance contributes the most by conducting feature decomposition on the covariance matrix of the data. The feature vector corresponding to the largest eigenvalue is then the main direction indicated by the data distribution and is, therefore, taken as the average axis of rotation.

### 4.1.2 Calibration of heading angle and displacement based on nonlinear optimization

Robot motion restricted to a horizontal plane incurs rotation excitation only about the Z axis of the **O** coordinate system. Therefore, the Z axis component $\mathbf{p}^O_{B_z}$ of the spatial position of the origin of the **B** coordinate system in the **O** coordinate system (i.e., $\mathbf{p}^O_B$) cannot be calibrated using rotation excitation and must be obtained by manual measurement. This yields the rotation matrix from the **B** coordinate system to the **O** coordinate system (i.e., $\mathbf{R}^O_B$). Then, the pose transformation $\mathbf{p}^O_F, \mathbf{R}^O_F$ between the **O** coordinate system and the **F** coordinate system must be calibrated. According to the definition of the **F** coordinate system, we know that $\mathbf{p}^O_F = \begin{bmatrix} \mathbf{p}^O_{B_x} & \mathbf{p}^O_{B_y} & 0 \end{bmatrix}^T$, and the pitch and roll angles of $\mathbf{R}^O_F$ are 0. Therefore, the transformation matrix from the **F** coordinate system to the **O** coordinate system (i.e., $\mathbf{T}^O_F$) includes three degrees of freedom, which are the X-axis position $\mathbf{p}^O_{F_x}$, the Y-axis position $\mathbf{p}^O_{F_y}$, and the heading angle $\mathbf{\theta}^O_F$.

We employ VIO and wheel speed inertial odometry to record the robot pose data over a specific period of time, and the data are saved as path files. Then, $\mathbf{p}^F_B$ and $\mathbf{R}^F_B$ are used to convert the VIO path to the **F** coordinate system. Because the IMU is fixed on the chassis, the relative pose of the VIO path between frames $\hat{\mathbf{p}}^{F_i}_{F_{i+1}}$ and the relative pose of the wheel speed inertial odometer $\hat{\mathbf{p}}^{O_i}_{O_{i+1}}$ have the following correspondences:

$$\left[\mathbf{p}^{O_i}_{O_{i+1}}\right]_{\text{VIO}} = \left[\mathbf{p}^{O_i}_{O_{i+1}}\right]_{\text{WheelOdometry}}$$

$$\mathbf{p}^O_F + \mathbf{R}\{\mathbf{\theta}^O_F\}\hat{\mathbf{p}}^{F_i}_{F_{i+1}} - \mathbf{R}^{O_i}_{O_{i+1}}\mathbf{p}^O_F = \begin{bmatrix} s_x^{-1} & 0 \\ 0 & s_y^{-1} \end{bmatrix}\hat{\mathbf{p}}^{O_i}_{O_{i+1}} \quad (10)$$

The difference between the left and right sides of (10) is used as the residual $\mathbf{r}_{i,i+1}$, and the chassis motion model parameters and the chassis–IMU external parameters are used as variables to define the following nonlinear least squares optimization problem:

$$\mathbf{x}^* = \underset{\mathbf{x}}{\operatorname{argmin}} \sum_{i=1}^{k-1} \rho \|\mathbf{r}_{i,i+1}(\mathbf{x})\|^2, \tag{11}$$

where $\mathbf{x} = \begin{bmatrix} \mathbf{p}_F^O & \theta_F^O & s_x & s_y \end{bmatrix}^T$ and $\mathbf{p}_F^O = \begin{bmatrix} \mathbf{p}_{F_x}^O & \mathbf{p}_{F_y}^O \end{bmatrix}^T$ are the vectors of variables to be optimized. The residual is defined as.

$$\begin{aligned}
\mathbf{r}_{i,i+1}(\mathbf{x}) &= \left[\mathbf{p}_{O_{i+1}}^{O_i}\right]_{\text{VIO}} - \left[\mathbf{p}_{O_{i+1}}^{O_i}\right]_{\text{WheelOdometry}} \\
&= (\mathbf{p}_{O_i \to F_i}^{O_i} + \mathbf{p}_{F_i \to F_{i+1}}^{O_i} + \mathbf{p}_{F_{i+1} \to O_{i+1}}^{O_i}) - \begin{bmatrix} s_x^{-1} & 0 \\ 0 & s_y^{-1} \end{bmatrix} \hat{\mathbf{p}}_{O_{i+1}}^{O_i} \\
&= (\mathbf{p}_F^O + \mathbf{R}\{\theta_F^O\}\hat{\mathbf{p}}_{F_{i+1}}^{F_i} - \mathbf{R}_{O_{i+1}}^{O_i}\mathbf{p}_{O_{i+1} \to F_{i+1}}^{O_{i+1}}) - \begin{bmatrix} s_x^{-1} & 0 \\ 0 & s_y^{-1} \end{bmatrix} \hat{\mathbf{p}}_{O_{i+1}}^{O_i} \\
&= \mathbf{p}_F^O + \mathbf{R}\{\theta_F^O\}\hat{\mathbf{p}}_{F_{i+1}}^{F_i} - \mathbf{R}_{O_{i+1}}^{O_i}\mathbf{p}_F^O - \begin{bmatrix} s_x^{-1} & 0 \\ 0 & s_y^{-1} \end{bmatrix} \hat{\mathbf{p}}_{O_{i+1}}^{O_i}
\end{aligned} \tag{12}$$

Here, $\mathbf{R}\{\theta\}$ is a 2D rotation matrix, and $\mathbf{R}\{\theta\} = \begin{bmatrix} \cos\theta & -\sin\theta \\ \sin\theta & \cos\theta \end{bmatrix}$.

The solution to the nonlinear least squares optimization problem in (11) requires the appropriate partial derivatives of (12). First, the partial derivative of the residual relative to the position increment $\delta\mathbf{p}_F^O$ is as follows:

$$\frac{\partial \mathbf{r}([\mathbf{p}_F^O + \delta\mathbf{p}_F^O \quad \theta_F^O \quad s_x \quad s_y]^T)}{\partial \delta\mathbf{p}_F^O} = \mathbf{I} - \mathbf{R}_{O_{i+1}}^{O_i}. \tag{13}$$

Based on the definition of $\mathbf{R}\{\theta\}$, this yields the following:

$$\frac{\partial [\mathbf{R}\{\theta + \delta\theta\}\mathbf{p}]}{\partial \delta\theta} = \begin{bmatrix} |\mathbf{p}|\cos(\theta + \theta_\mathbf{p} + \pi/2) \\ |\mathbf{p}|\sin(\theta + \theta_\mathbf{p} + \pi/2) \end{bmatrix}, \tag{14}$$

where $\theta_\mathbf{p} = \operatorname{atan}(p_y, p_x)$. Therefore, the partial derivative of the residual relative to the heading angle increment $\delta\theta_F^O$ is given as follows:

$$\begin{aligned}
\frac{\partial \mathbf{r}([\mathbf{p}_F^O + \delta\mathbf{p}_F^O \quad \theta_F^O \quad s_x \quad s_y]^T)}{\partial \delta\mathbf{p}_F^O} &= \frac{\partial [\mathbf{R}\{\theta_F^O + \delta\theta_F^O\}\hat{\mathbf{p}}_{F_{i+1}}^{F_i}]}{\partial \delta\theta_F^O} \\
&= \begin{bmatrix} |\hat{\mathbf{p}}_{F_{i+1}}^{F_i}|\cos(\theta_F^O + \theta_{\hat{\mathbf{p}}_{F_{i+1}}^{F_i}} + \pi/2) \\ |\hat{\mathbf{p}}_{F_{i+1}}^{F_i}|\sin(\theta_F^O + \theta_{\hat{\mathbf{p}}_{F_{i+1}}^{F_i}} + \pi/2) \end{bmatrix}.
\end{aligned} \tag{15}$$

The partial derivative of the residual relative to the parameter increment $\delta s_x$ of the chassis motion model is as follows:

$$\frac{\partial \mathbf{r}([\mathbf{p}_F^O \quad \theta_F^O \quad s_x + \delta s_x \quad s_y]^T)}{\partial \delta(s_x^{-1})} = \lim_{\delta s_x \to 0} \frac{\left(\begin{bmatrix} s_x + \delta s_x & 0 \\ 0 & s_y \end{bmatrix} \hat{\mathbf{p}}_{O_{i+1}}^{O_i}\right) - \left(-\begin{bmatrix} s_x & 0 \\ 0 & s_y \end{bmatrix} \hat{\mathbf{p}}_{O_{i+1}}^{O_i}\right)}{\delta s_x} = \begin{bmatrix} -1 & 0 \\ 0 & 0 \end{bmatrix} \hat{\mathbf{p}}_{O_{i+1}}^{O_i}. \tag{16}$$

Similarly, the partial derivative of the residual relative to the parameter increment $\delta s_y$ of the chassis motion model is as follows:

$$\frac{\partial \mathbf{r}([\mathbf{p}_F^O \quad \theta_F^O \quad s_x \quad s_y + \delta s_y]^T)}{\partial \delta(s_y^{-1})} = \begin{bmatrix} 0 & 0 \\ 0 & -1 \end{bmatrix} \hat{\mathbf{p}}_{O_{i+1}}^{O_i}. \tag{17}$$

The solution to the nonlinear least squares optimization problem in (11) provides the chassis–IMU external parameters.

### 4.2 IMU calibration

#### 4.2.1 Calibration of systematic errors

The process of calibrating the systematic errors of the IMU is divided into the following steps:

① Hold the IMU motionless for 50 s to estimate the zero offset error of the gyroscope.

② Rotate the IMU by a specified angle about a given axis of the gyroscope and maintain that position for 5

s.
③ Repeat step ② for all three axes of the gyroscope for a total of six rotational directions, and for all of the three axes of the accelerometer for a total of six directions.
④ The recorded sensor data are converted into a suitable format, and $\mathbf{T}_a, \mathbf{K}_a, \mathbf{b}_a, \mathbf{T}_g, \mathbf{K}_g$, and $\mathbf{b}_g$ are calibrated using previously developed open source tools [11].

The calibration results obtained for the BMI055 IMU are listed in Table 1. Accordingly, the calculated accelerometer axis deviation is as follows:

$$\begin{bmatrix} 1.0 & -0.0388 & -0.0025 \\ 0.0 & 1.0 & 0.0223 \\ 0.0 & 0.0 & 1.0 \end{bmatrix},$$

while the calculated gyroscope axis deviation is as follows:

$$\begin{bmatrix} 1.0 & -0.0573 & 0.00110 \\ 0.0647 & 1.0 & 0.01660 \\ 0.0038 & -0.0150 & 1.0 \end{bmatrix}.$$

Table 1. Systematic errors of the BMI055 inertial measurement unit

|  |  | Accelerometer | Gyroscope |
| --- | --- | --- | --- |
| Zero offset | X axis | 0.080551 m/s$^2$ | −0.0032665 rad/s |
|  | X axis | 0.119632 m/s$^2$ | −0.0044932 rad/s |
|  | Y axis | −0.340042 m/s$^2$ | 0.0010749 rad/s |
| Scale factor | X axis | 1.01807 | 0.99514 |
|  | Y axis | 1.01469 | 1.00125 |
|  | Z axis | 1.00625 | 0.99586 |

**4.2.2 Calibration of random errors**

The process of calibrating the random errors of the IMU is divided into the following steps:
① Fix the IMU on a stable plane to avoid the impact of vibration on the calibration accuracy. The IMU is also fixed in the same orientation as it is on the robot chassis with the chassis positioned on a level surface to reduce the impact of the acceleration due to gravity on gyroscope measurements.
② Connect the IMU to the main control board through an extension cable, and establish communication between the main control board and a personal computing device.
③ Wait for several minutes until the system enters a stable state.
④ Record IMU data for 5 to 10 h using the rosbag tool of the open source Robot Operating System (ROS) while holding the IMU stationary.
⑤ Use the ROS imu_utils tool to modify the sampling frequency of the IMU in the source code to the selected value, and then fit the data to the variance trends, as discussed in Subsection 3.1.2, to obtain the calibration results.

The calibration results of the BMI055 IMU are listed in Table 2. We note that the calibration results are basically consistent with the values given in the BMI055 data sheet.

Table 2. Random errors of the BMI055 inertial measurement unit

|  |  | White noise | Bias instability |
| --- | --- | --- | --- |
| Accelerometer | X axis | $2.938 \times 10^{-3}$ rad/s | $1.352 \times 10^{-5}$ rad/s$^2$ |
|  | Y axis | $4.813 \times 10^{-3}$ rad/s | $1.085 \times 10^{-5}$ rad/s$^2$ |
|  | Z axis | $6.184 \times 10^{-3}$ rad/s | $1.920 \times 10^{-5}$ rad/s$^2$ |
| Gyroscope | X axis | $1.103 \times 10^{-1}$ m/s$^2$ | $1.194 \times 10e^{-3}$ m/s$^3$ |
|  | Y axis | $2.980 \times 10^{-2}$ m/s$^2$ | $1.996 \times 10^{-4}$ m/s$^3$ |

| | Z axis | $3.271 \times 10^{-2}$ m/s$^2$ | $2.904 \times 10^{-4}$ m/s$^3$ |

## 4.3 Camera calibration

### 4.3.1 Calibration of model parameters

The open source Kalibr toolbox was employed to calibrate the camera model parameters. As discussed in Subsection 3.2, the pinhole camera model in (6) was applied for the RGB camera, while the universal camera model was applied for the fisheye camera. The specific calibration steps are given as follows:

① Prepare a non-reflective checkerboard calibration board, and record the parameters of the calibration board into the configuration file of the Kalibr toolbox.
② Use the camera to capture images of the calibration board in various positions of the camera's field of view. Both the camera and the calibration board must be kept stable during the image capture process.
③ Save the captured images in the bag file of the Kalibr toolbox.
④ Obtain the parameters of the specified camera projection model using the kalibr_calibrate_cameras tool in the Kalibr toolbox.

The internal calibration results of the RGB and fisheye camera components of the Intel RealSense ZR300 depth camera are listed in Table 3. The re-projection errors of the RGB and fisheye cameras after calibration were both less than 0.5 pixels, indicating that the calibration results were good.

Table 3. Intel RealSense ZR300 depth camera internal calibration parameters

| Parameter | RGB camera | Fisheye camera |
|---|---|---|
| $f_x$ | 617.92 | 761.95 |
| $f_y$ | 618.54 | 761.42 |
| $c_x$ | 316.07 | 309.99 |
| $c_y$ | 244.96 | 234.27 |
| $k_1$ | 0.1182 | −0.07772 |
| $k_2$ | −0.2507 | 0.2731 |
| $p_1$ | $-4.410 \times 10^{-4}$ | $-2.380 \times 10^{-3}$ |
| $p_2$ | $2.824 \times 10^{-4}$ | $3.120 \times 10^{-3}$ |
| $\xi$ | 1 | 1.743 |

### 4.3.2 Camera-IMU external calibration

The Kalibr toolbox was again used for the calibration of the camera-IMU external parameters [14]. The specific calibration process is given as follows:

① Identify the calibration plate in each image frame based on the known calibration plate.
② Apply global structure from motion (SfM) to obtain the camera pose in each image frame according to the identified corner points of the calibration plate.
③ Calibrate the rotation matrix $\mathbf{R}_C^B$ between the **C** and **B** coordinate systems. This can be conducted through rotation excitation because the camera and the IMU are rigidly connected.
④ After determining $\mathbf{R}_C^B$, the displacement between the respective origins of these coordinate systems $\mathbf{p}_C^B$ can be obtained by finding the least squares solution of the overdetermined equations.

The calibration results yielded the following transformation matrix:

$$\mathbf{T}_C^B = \begin{bmatrix} 0.9991 & -0.0395 & 0.0124 & 0.097 \\ 0.0393 & 0.9992 & 0.0098 & 0.0084 \\ -0.0128 & -0.0093 & 0.9999 & -0.0002 \\ 0 & 0 & 0 & 1 \end{bmatrix}.$$

## 4.4 Calibration of chassis−IMU internal and external parameters

### 4.4.1 Calibration of pitch and roll angles

The process of employing PCA for obtaining principal rotation directions in all rotation measurements is

divided into the following steps:

Generate the PCA data set composed of $K$ elements:

$$\mathbf{X} = [\widehat{\mathbf{w}}_0^B \quad -\widehat{\mathbf{w}}_0^B \quad \widehat{\mathbf{w}}_1^B \quad -\widehat{\mathbf{w}}_1^B \quad \cdots \quad \widehat{\mathbf{w}}_K^B \quad -\widehat{\mathbf{w}}_K^B]^T, \tag{18}$$

where $\widehat{\mathbf{w}}_i^{BT}$ is the angular velocity measurement of the $i$-th frame in the **B** coordinate system. The reverse angular velocity measurement is added to the data set to ensure that the average value of the data is 0.

Calculate the covariance matrix of $\mathbf{X}$ as follows: $\Sigma = \frac{\mathbf{X}^T \mathbf{X}}{K-1}$.

Perform eigen decomposition of the covariance matrix to obtain the eigenvalues $\mathbf{\Lambda} = [\lambda_1 \quad \lambda_2 \quad \lambda_3]^T$ and the corresponding feature vector: $\mathbf{V} = [\mathbf{v}_1 \quad \mathbf{v}_2 \quad \mathbf{v}_3]^T$.

Select the largest eigenvalue $\lambda_{max}$ and its eigenvector $\mathbf{v}_{max}$, where $\mathbf{v}_{max}$ is the average rotation direction in the **B** coordinate system. The direction of the Z axis in the **O** coordinate system is then either $\mathbf{v}_{max}$ or $-\mathbf{v}_{max}$, where, according to the pre-estimated relative rotation matrix $\mathbf{R}_B^O$, this is given as $\mathbf{v}_{max}$ if $\mathbf{R}_B^O \mathbf{v}_{max} \cdot [0 \quad 0 \quad 1]^T > 0$, and is given as $-\mathbf{v}_{max}$ otherwise.

The origin of the **F** coordinate system in the **O** coordinate system can be defined as $\mathbf{p}_F^O = [\mathbf{p}_{Bx}^O \quad \mathbf{p}_{By}^O \quad 0]^T$, such that $\mathbf{p}_B^F = [0 \quad 0 \quad \mathbf{p}_{Bz}^O]^T$. The pitch and roll angles of $\mathbf{R}_F^O$ are both 0, and the heading angle is the same as that of $\mathbf{R}_B^O$. Meanwhile, the heading angle of $\mathbf{R}_B^F$ is 0, and the pitch and roll angles are the same as those of $\mathbf{R}_B^O$. In addition, the relationship $\boldsymbol{\theta}_F^B = \frac{\mathbf{v}_{max}}{|\mathbf{v}_{max}|} \times [0 \quad 0 \quad 1]^T$ can be derived from $\mathbf{R}_F^B \frac{\mathbf{v}_{max}}{|\mathbf{v}_{max}|} = [0 \quad 0 \quad 1]^T$. Therefore, the pitch and roll angles between the coordinate systems are given as follows:

$$\begin{bmatrix} 0 \\ pitch_B^F \\ roll_B^F \end{bmatrix} = \text{YPR}(\boldsymbol{\theta}_F^B) \tag{19}$$

Based on actual experimental data, the pitch angle of the **B** coordinate system relative to the **O** coordinate system for the example mobile robot implementation was −2.7373° and the corresponding roll angle was −91.1430°.

### 4.4.2 Calibration of heading angle, displacement, and chassis internal parameters

The chassis–IMU external parameters and the chassis internal parameters were obtained by solving the nonlinear least squares optimization problem in (11) using the Google Ceres non-linear optimization library **[15]**. The calibration results obtained from three calibration experiments are listed in Table 4. The average of the three calibration results is employed as the final calibration result.

Table 4. Calibration results of chassis–inertial measurement unit (IMU) external parameters and chassis internal parameters

|  | Chassis–IMU external parameters | | | Chassis internal parameters | |
| --- | --- | --- | --- | --- | --- |
|  | $\mathbf{p}_{F_x}^O$ (m) | $\mathbf{p}_{F_y}^O$ (m) | $\theta_F^O$ (°) | $s_x^{-1}$ (%) | $s_y^{-1}$ (%) |
| Experiment 1 | 0.087 | 0.066 | −89.42 | 101.63 | 102.84 |
| Experiment 2 | 0.110 | 0.066 | −89.28 | 99.838 | 104.64 |
| Experiment 3 | 0.1042 | 0.058 | −89.16 | 97.725 | 103.75 |
| Standard deviation | 0.0121 | 0.004 | 0.128 | 1.957 | 0.900 |
| Mean | 0.1008 | 0.064 | −89.29 | 99.733 | 103.74 |

## 5. Conclusion

This study presented a novel chassis–IMU internal and external parameter calibration algorithm based on nonlinear optimization, which was designed for robots equipped with cameras, IMUs, and wheel speed odometers. All of the internal and external reference calibrations in the algorithm are conducted using the robot's existing equipment without the need for additional calibration aids. The proposed calibration process is simplified by decoupling the parameter calibration into two main steps: the first step is employed when robot motion is restricted to the horizontal plane, which limits rotations to occur only about the Z axis of the chassis. Then, the pitch and roll angles of the IMU in the chassis coordinate system are obtained by calculating the direction of the rotation axis in the IMU coordinate system using PCA. The second step employs pose data records of the robot obtained over a specific period using VIO and wheel speed inertial odometry. The relationship between the relative poses between successive time frames is employed as pose data with which a nonlinear least squares problem is defined for optimizing both the interior parameters of the chassis and chassis−IMU external parameters. The feasibility of the method was verified by its application to a Mecanum wheel omnidirectional mobile platform as an example. The example demonstrated the ease with which the proposed calibration algorithm is implemented, and the process is shown to guarantee the accuracy of calibration.

## ACKNOWLEDGMENT

The work of paper was supported by National Natural Science Foundation of China(No.61672244, No. 91748106), Hubei Province Natural Science Foundation of China(No. 2019CFB526), and Key Technology Project of China Southern Power Grid(GZKJQQ00000164).

**About the Author**

**Gang Peng**, received the doctoral degree from the Department of control science and engineering of Huazhong University of Science and Technology (HUST) in 2002. Currently, he is an associate professor in the Department of Automatic Control, School of Artificial Intelligence and Automation, HUST. He is also a senior member of the China Embedded System Industry Alliance and the China Software Industry Embedded System Association, a senior member of the Chinese Electronics Association, and a member of the Intelligent Robot Professional Committee of Chinese Association for Artificial Intelligence. His research interests include intelligent robots, machine vision, multi-sensor fusion, machine learning and artificial intelligence.

**Zezao Lu**, received bachelor degree in school of automation from Center South University, China, in 2016. He received master degree at the Department of Automatic Control, School of Artificial Intelligence and Automation, Huazhong University of Science and Technology. His research interests are intelligent robots and perception algorithms.

**Zejie Tan**, received bachelor degree in school of automation from Wuhan University of Science and Technology, China, in 2019. He is currently a graduate student at the Department of Automatic Control, School of Artificial Intelligence and Automation, Huazhong University of Science and Technology. His research interests are intelligent robots and perception algorithms.

**He Dingxin**, received master degree in the department of automatic control of Huazhong University of Science and Technology (HUST) in 1995. Currently, he is a professor in the Department of Automatic Control, School of Artificial Intelligence and Automation, HUST. His research interests include intelligent robots, embedded system and artificial intelligence.

**Xinde Li**, professor and doctoral tutor, graduated from the Control Department of Huazhong University of Science and Technology in June 2007, worked for the School of Automation of Southeast University in December of the same year. From January 2012 to January 2013 as a national public visiting scholar at Georgia Tech Visit and exchange for one year. He was selected as an IEEE Senior member in 2016. From January 2016 to the end of August 2016, he worked as a Research Fellow in the ECE Department of the National University of Singapore. His research interests include intelligent robots, machine vision perception, machine learning, human-computer interaction, intelligent information fusion and artificial intelligence.